\documentclass[10pt,twocolumn,letterpaper]{article}

\usepackage{iccv}
\usepackage{times}
\usepackage{epsfig}
\usepackage{graphicx}
\usepackage{amsmath}
\usepackage{amssymb}
\usepackage{amssymb}
\usepackage{appendix}
\usepackage{caption}
\usepackage{float}
\usepackage{bm}
\usepackage{stfloats}
\iccvfinalcopy

\begin{document}

%%%%%%%%% TITLE

\title{Integrating neural networks into the blind deblurring framework to compete with end-to-end learning based methods}

\author{Junde Wu \and Xiaoguang Di\and Jiehao Huang \and Yu Zhang \\
Control \& Simulation Center, Harbin Institution of Technology\\
{\tt\small wujunde@hit.edu.cn, dixiaoguang@hit.edu.cn}
}

\maketitle

\def\iccvPaperID{2342} % *** Enter the ICCV Paper ID here
\def\httilde{\mbox{\tt\raisebox{-.5ex}{\symbol{126}}}}

% Pages are numbered in submission mode, and unnumbered in camera-ready
\ificcvfinal\pagestyle{empty}\fi
\setcounter{page}{4321}

%%%%%%%%% ABSTRACT
\begin{abstract}
  Recently, end-to-end learning methods based on deep neural network (DNN) have been proven effective for blind deblurring. Without human-made assumptions and numerical algorithms, they are able to restore images with fewer artifacts and better perceptual quality. However, in practice, we also find some of their drawbacks. Without the theoretical guidance, these methods can't perform well when the motion is complex and sometimes generate unreasonable results. In this paper,  for overcoming these drawbacks, we integrate deep convolution neural networks into conventional deblurring framework. Specifically, we build Stacked Estimate Residual Net (SEN) to estimate the motion flow map and Recurrent Prior Generative and Adversarial Net (RP-GAN) to learn the implicit image prior in the optimization model. Comparing with state-of-the-art end-to-end learning based methods, our method restores reasonable details and shows better generalization ability.
\end{abstract}
%%%%%%%%% BODY TEXT
\section{Introduction}

Motion blur is a commonly appeared degradation of  image qualities. The blurry images generally caused by the shakes of camera and fast object motions.
%Most of the deblurring methods are modelled by :
%\begin{equation}\label{equation:first}
%O = I_{s}*K + n
%\end{equation}
%\noindent
%where * denotes convolution operator, \(O\), \(I_{s}\), \(K\), \(n\) are observed blurry image, latent sharp image, blur kernel and noise respectively. 
The problem is highly ill-posed due to the unknown blur kernel and extra noise.
 \begin{figure}[h]
 \centering
  \includegraphics[width = 0.5\textwidth]{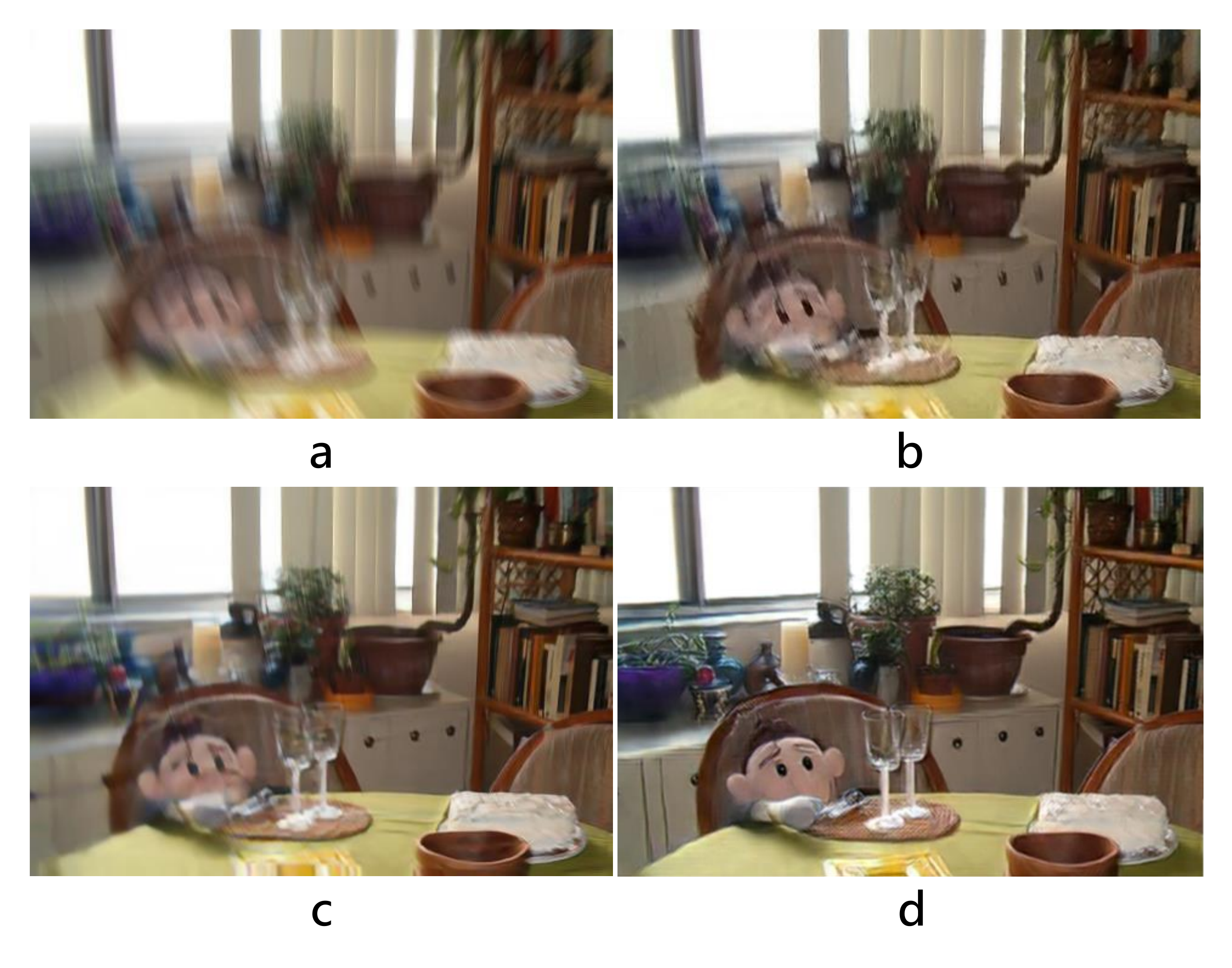}
  \caption{A deblurred example. (a) Blurred image. (b) Result of Nah \textit{et al}. \cite{Nah}. (c) Result of Tao \textit{et al}. \cite{Tao2018Scale}. (d) Ours.}
  \label{show}
\end{figure}

The end-to-end learning based methods solve it by training a neural network to restore the clear images from blurred observations directly. Depending on  the strong fitting ability of deep neural network (DNN), they can restore images with fewer artifacts and better visual effect. Previous work \cite{Nimisha2017Blur,Nah,tao2018srndeblur,DeblurGAN} had attempted various kinds of neural networks, including Residual Network (ResNet), Multi-scale Convolutional Neural Network and Generative and Adversarial Net (GAN). These work proved deep and well-designed network structure is the key to success. However, in practice, we also find some limitations of this kind of methods. Firstly, most of them can't perform well when the motion is too large or the blur is highly space-variant. Besides, sometimes they restore heavy blurry content to unreasonable objects since lacking theoretical guidance. The relevant results are shown in our experiment.

On the other side, conventional deblurring methods generally focus on getting the better blur kernel and image prior. The development of the neural network encourages many attempts to embed the neural network into the conventional deblurring framework. Utilizing the neural networks to estimate the blur kernel seems to be straightforward and effective \cite{Sun2015Learning,Dong2017From,Schuler2016Learning,Yan2016Blind}, while learning image prior is somewhat challenging. One way to do so is learning the implicit image prior directly from ground truth \cite{Zhang2017Learning,dong2018denoising,Schuler2013A,burger2012image}. A typical strategy of this kind of methods is using half-quadratic splitting algorithm to split the non-blind restoration task to two sub-problems and then alternatively solve them to get the final solution. The deconvolution sub-problem can be solved by numerical method directly. Another denoise sub-problem, which contains the unknown image prior, is solved by the neural network. However, it is not easy to build such a neural network which can adapt to the iterative optimization process well. Zhang \textit{et al}. \cite{Zhang2017Learning} pre-trained multiple CNN for image denoising task and integrated them into the optimization-based framework. However, these pre-trained parameters can not dynamically adjust with the iterative optimization. Dong \textit{et al}. \cite{dong2018denoising} computed the inexact solution of the deconvolution sub-problem with a single step of gradient descent. Although it enables the whole optimization model to be trainable from end to end, it also takes extra optimization steps. Moreover, since both of them \cite{Zhang2017Learning,dong2018denoising} are trained for non-blind restoration, they are hard to be applied to blind deblurring directly.\\
\begin{figure*}[ht]
\centering
\includegraphics[width = \textwidth]{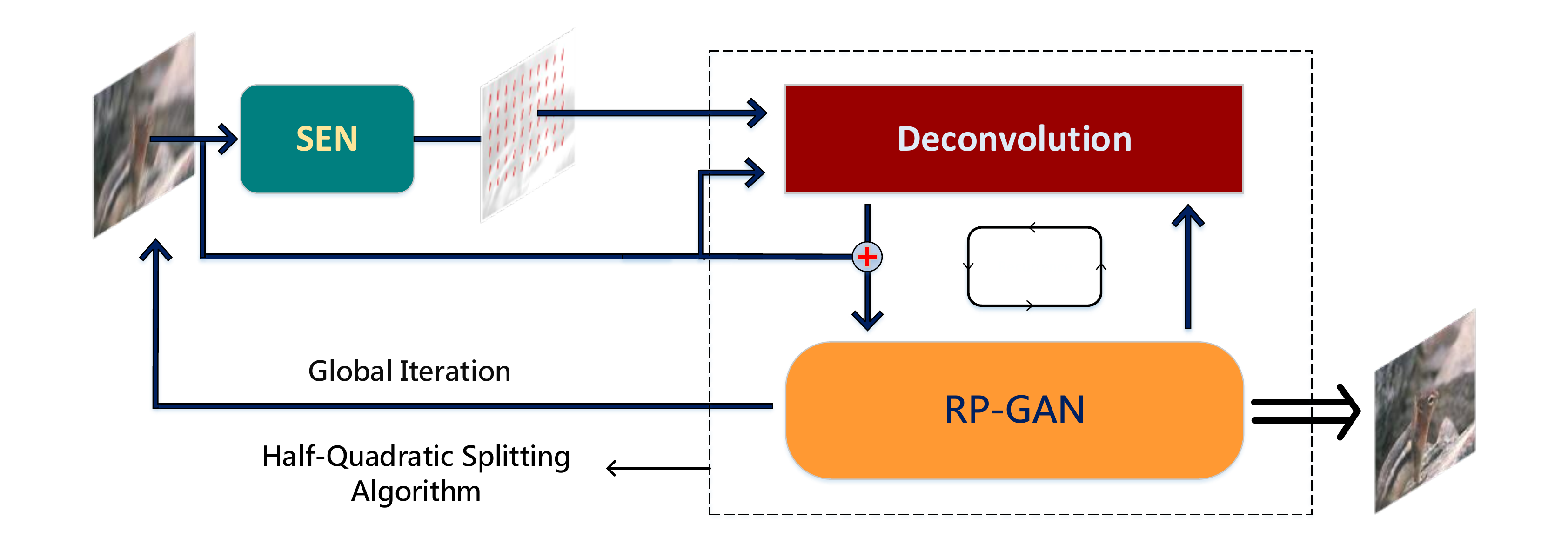}
\caption{The deblurring system}
    \includegraphics[scale = 0.1]{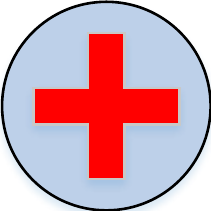} denotes concatenate. \includegraphics[scale = 0.1]{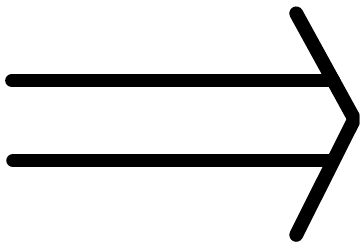} denotes system output. The blurred image is firstly sent to SEN for estimating the motion flow map. Then the blurred image can be deconvolved by the estimated motion flow map and then send to RP-GAN's recurrent process. The result of RP-GAN will serve as the blurred image and be sent to the system again in global iteration.
\label{fig:system}
\end{figure*}
We can see either the end-to-end learning methods or the optimization-based methods have their drawbacks. However, end-to-end learning deblurring methods are more often applied in practice recently considering their higher comprehensive competence---their better visual effects, higher speed and complete blind deblurring models which make them can be conveniently used. But in this paper, we find optimization-based methods have the potential to outperform the end-to-end learning based methods and make up for their shortcomings. We design our optimization part following the implicit-prior-learning strategy, but we build a Generative and Adversarial Net(GAN) with deeper layers and higher ability to generate more realistic images. We introduce the recurrent structure in the network and enable it can be trained online with the iterative optimization process. We call the network as Recurrent Prior Generative and Adversarial Net (RP-GAN). For blind deblurring, we build a stacked residual network, called Stacked Estimate Net (SEN), to estimate the space-varying motion flow of the blurred image. Although the estimation strategy is straightforward, its results are the best comparing with other similar methods. Furthermore, in order to ensure the robustness of RP-GAN to the inexact blur kernels, we add an extra discriminator to impose extra punishment on artifacts caused by the incorrectly-estimated kernels. In the end, we send the result to the whole system again to further process the residue blur in global iteration. Our intact blind deblurring model is shown in Figure \ref{fig:system}. To our knowledge, this is the first optimization-based deblurring method comparing with the powerful end-to-end learning based methods in the experiment and getting the better performance.

In conclusion, the main contributions of this paper are:
\begin{itemize}
\item We build SEN to estimate the nonuniform motion flow map. Its accuracy exceed all the other deep learning based methods.
\item We propose RP-GAN and integrate it into the iterative optimization-based framework well. 
\item We design the aforementioned two parts to be well-compatible, thus making the integrated blind deblurring model to be robust and achieve better deblurring effect. 
\end{itemize}

%-------------------------------------------------------------------------
\section{Related Work}
Most conventional optimization-based deblurring approaches get success depending on guiding maximum a posteriori probability (MAP) process by assumed priors, such as the total variational regularizer \cite{TV1,TV2}, Gaussian scale mixture priors \cite{Fergus2006Removing}, normalized sparsity \cite{Krishnan2011Blind}, L0 gradients \cite{Xu2013Unnatural}, dark channel prior \cite{darkchannel}, \textit{etc}. However, since all these hand-crafted priors are designed under limited observations or restricted assumptions, these algorithms show unsatisfied results when processing the images that have fewer corresponding features.

The development of neural networks inspires many works to utilize it in conventional deblurring framework. Some work focus on learning the blur kernel \cite{Schuler2016Learning, Yan2016Blind,Sun2015Learning,Dong2017From}. Sun \textit{et al}. \cite{Sun2015Learning} proposed a convolutional neural network (CNN) to predict the probabilistic distribution of blur. Gong \textit{et al}. \cite{Dong2017From} proposed a fully-convolutional deep neural network (FCN) to directly estimate the blur kernel in pixel level and get a higher accuracy. On the other side, several work learn generic image priors (explicit or implicit) for optimization \cite{Sreehari2016Plug,Zhang2017Learning,Schuler2013A,Zhang2016Learning,er,bigdeli2017image,dong2018denoising,burger2012image}. In terms of learning explicit prior, Zhang \textit{et al}. \cite{Zhang2016Learning} learned denoised gradient as image prior to guide the deconvolution model. Li \textit{et al}. \cite{er} learned a discriminative prior for deblurring in generic scenarios. However, they have to compromise on restoring quality when embedding an explicit prior to numerical methods. As for learning implicit prior, Zhang \textit{et al}. \cite{Zhang2017Learning} used pre-trained multiple CNN to restore the corrupted deconvolutional results. Dong \textit{et al}. \cite{dong2018denoising} made a compromise on deconvolutional solutions to make the whole optimization model can be trained from end to end. Nevertheless, these operations are lossy. Besides, these methods are hard to be applied to blind deblurring directly. Furthermore, their networks are actually too simple to fit their tasks. 

Recently, end-to-end learning based methods have shown great advantages for deblurring. Nah \textit{et al}. \cite{Nah} proposed a multi-scale CNN for restoring the images in three different levels. Each level processes different blur scale, from small to large. Kupyn \textit{et al}. \cite{DeblurGAN} built GAN model for deblurring directly from the blurred observations to the sharp images. Tao \textit{et al}. \cite{tao2018srndeblur} inherited the multi-scale structure from Nah \textit{et al}. \cite{Nah} and added long short-term memory (LSTM) in the recurrent process. End-to-end learning based methods restore the images with fewer artifacts than optimization-based methods. But they are highly dependent on the observed data. Thus, on the one hand, they often show poor generalization ability, especially when the motion is large or complex. On the other hand, since these methods completely discard the deblurring principle, they sometimes restore unreasonable contents from blurred images, which may cause the restored images to be more confusing than the blurred ones.
\section{Motion Flow Estimation by SEN}
We propose a stacked residual network to estimate spatially-varying motion flow map from end to end. Our motion flow map is modeled following \cite{Dong2017From}. Specifically, given an arbitrary RGB blurry image \(O\) which has the size \(H*W\), the task is generating a  \(H*W*2\) matrix \(U\) to represent motion flow map.  \(U\) can be expressed as : \(U(i,j,1) = u_{i,j}\) , \(U(i,j,2) = v_{i,j}\), \(\forall i,j \in O\). \(u\) and \(v\) denote horizontal and vertical motions respectively. 

The network we propose, called SEN, is mainly based on an encoder-decoder structure which has been proven useful in various vision tasks \cite{huang23,huang31,huang33,huang39,hourglass}. We adapt the network structure from \cite{hourglass}. The detailed structure is shown in Appendix A. Unlike the symmetrical structure used in \cite{hourglass}, we use three dilated convolution layers with increasing channels  in front of the network. They help to enlarge the receptive field and abstract deep blur features .

%Because the prior layers are more likely related to the image content but not the blur features. Skip connections are added only after dilated convolution layers. From experiments, we find the skip connections on prior layers impair the network performance because of the large difference between the image content and the motion flow map. 

We also use stacked structure to improve the estimation accuracy. Stacked structure takes the result of the last network as the input of the next one. In the paper, the stacked networks share the similar structures. The result from the first network will be concatenated with the blurred image to serve as the next network's input. Although more stacks can further improve the estimation accuracy, they will also cause over-fitting. For the balance, we take two stacks in practice. 

In previous work \cite{Sun2015Learning,Dong2017From}, the task was treated as a classification problem. However, we find using regression loss can reach higher accuracy.
Specifically, L2 loss is used for training, which can be expressed as :
\begin{equation}\label{equation:mse}
\mathcal{L}_{L_{2}} = \frac{1}{N}\Vert U^{*} - U^{l} \Vert^{2}_{2}
\end{equation}
\noindent
where \(U^{*}\) denotes the estimated matrix and \(U^{l}\) denotes the label. \(N\) is the number of elements in the matrix. 
%-------------------------------------------------------------------------
\section{RP-GAN embedded in HQ Algorithm }
%In this section we introduce how we build a specific network and embed it in Half-Quadratic Splitting Algorithm(HQ Algorithm). The first part introduces the  HQ Algorithm. Second part shows the main features of RP-GAN and the insights behind it.
\subsection{Half-Quadratic Splitting Algorithm(HQ Algorithm)}
Once knowing the motion flow map, the deblurring issue can be modeled as:
\begin{equation}\label{equation:intact}
I = \arg min_{I} \Vert I * K - O\Vert^{2}_{2}+ \gamma \space p(I) 
\end{equation}
where \(I\) denotes the latent sharp image. \(O\) denotes the given blurry image. \(K\) denotes the heterogeneous motion blur kernel map with different blur kernels for each pixel in \(O\). \(p(I)\) denotes the latent prior of the image. \(*\) implies general convolution operation. \(\gamma\) is a weighting constant. Each nonuniform blur kernel in map \(K\) will be applied to its corresponding pixel. The specific operation is following \cite{Dong2017From}. In Eqn. (\ref{equation:intact}), since both \(I\) and \(p\) are unknown, it's hard to solve the equation directly.

Half-quadratic splitting algorithm is a way to simplify it. By HQS, we can split Eqn. (\ref{equation:intact}) into two sub-problems which show in Eqn. (\ref{equation:HQ}).
\begin{equation}\label{equation:HQ}
\left\{\begin{array}{ll}
\begin{split}
\ I^{*}_{n+1} = &\arg min_{I^{*}_{n+1}} \frac{\beta}{2}\Vert I^{*}_{n+1} - Z_{n}\Vert^{2}_{2} \\&+\frac{1}{2}\Vert I^{*}_{n+1}*K - O\Vert^{2}_{2} & {\raisebox{.5pt}{\textcircled{\raisebox{-.9pt} {1}}}}
\\
\ Z_{n+1}= &\arg min_{Z_{n+1}} \frac{\beta}{2}\Vert I^{*}_{n+1} - Z_{n+1}\Vert^{2}_{2} 
\\&+ \gamma \space p(Z_{n+1}) & {\raisebox{.5pt}{\textcircled{\raisebox{-.9pt} {2}}}}
\end{split}
\\
\end{array}
\right.
\end{equation}
where \(n\) is the number of iterations. \(I^{*}\) is the corrupted deconvolutional image, \(Z\) is an auxiliary variable initialized with observed blurry image \(O\). \(\beta\) is a variable parameter. Eqn. (\ref{equation:intact}) can be solved by alternatively solving two sub-problems with increasing \(\beta\). In Eqn. (\ref{equation:HQ}), Eqn. (\ref{equation:HQ})-\(\raisebox{.5pt}{\textcircled{\raisebox{-.9pt} {1}}}\) has the analytic solution :
\begin{equation}\label{equation:decon}
I^{*}_{n+1} = [K^{T}K + \beta\,\mathcal{I} ]^{-1}\, [\beta Z_{n} + K^{T} O]
\end{equation}                                                                                       
where \(\mathcal{I} \) is the identity matrix. But generally, the inverse matrix in it is hard to be computed directly. In practice, we use conjugate gradient algorithm to solve it.

Eqn. (\ref{equation:HQ})-\(\raisebox{.5pt}{\textcircled{\raisebox{-.9pt} {2}}}\) is a denoising module containing the unknown image prior, thus can not be solved without any assumption. Therefore, training a neural network to solve Eqn. (\ref{equation:HQ})-\(\raisebox{.5pt}{\textcircled{\raisebox{-.9pt} {2}}}\) is an appealing idea. In this way, we can learn the implicit image prior directly through data. 
\subsection{Recurrent Prior GAN and Insights behind it}
\begin{figure}[h]
\centering
  \includegraphics[scale = 0.6]{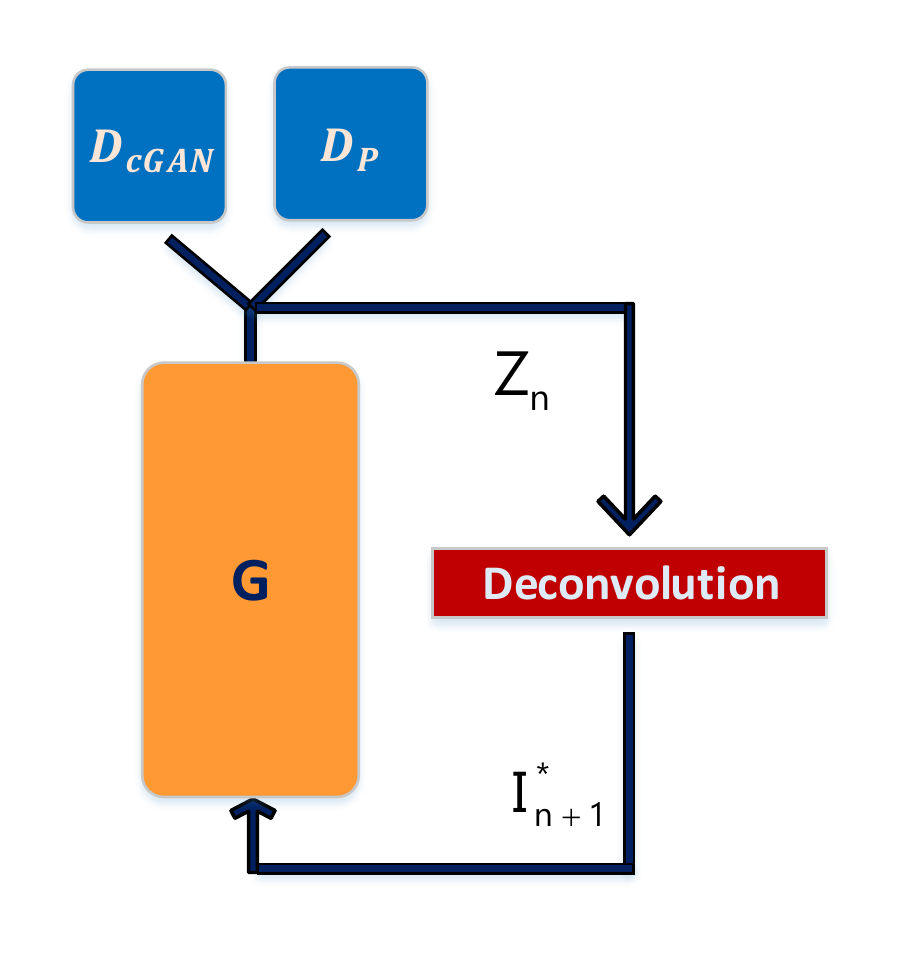}
  \caption{The recurrent process of RP-GAN. \(G\) denotes generator, \(D_{cGAN}\)  and \(D_{p}\) are cGAN discriminator and artifacts-penalize discriminator respectively}
  \label{fig:train_gan}
\end{figure}

For using neural networks to solve Eqn. (\ref{equation:HQ})-\(\raisebox{.5pt}{\textcircled{\raisebox{-.9pt} {2}}}\)  in different iterative steps, a straightforward way is training multiple neural networks to solve each of them separately \cite{Zhang2017Learning}. However, \cite{dong2018denoising} has explored the possibility that using one deep neural network to fit all these different modules. This point also matches our observation: a recurrent network will output clearer images when accepting better-quality deconvolutional results, which makes us believe network could learn to adjust the latent parameter \(\beta\) by itself (through perceiving variant inputs). So we build Recurrent Prior GAN (RP-GAN) to learn Eqn. (\ref{equation:HQ})-\(\raisebox{.5pt}{\textcircled{\raisebox{-.9pt} {2}}}\) from end to end. The basic structure of RP-GAN is roughly based on  \cite{imagetoimage}. We provide the details of it in Appendix A. In recurrent process, the restored result of the \(n\) recurrence (level) is served  as variable \(Z_{n}\)  for Eqn. (\ref{equation:HQ})-\(\raisebox{.5pt}{\textcircled{\raisebox{-.9pt} {1}}}\), then it will be used to calculate the corrupted deconvolutional value \(I^{*}_{n+1}\)  through Eqn. (\ref{equation:HQ})-\(\raisebox{.5pt}{\textcircled{\raisebox{-.9pt} {1}}}\). \(I^{*}_{n+1}\) will then be the input of \(n+1\) level in the recurrent generator. The process is shown in Figure \ref{fig:train_gan}. In the implementation, such a process will run three times. In another word, RP-GAN contains three levels.

However, although building one network with recurrent structure can prevent over-fitting, it also raises another problem: the training process will suffer heavy oscillation since the variance of RP-GAN will cause the change of its own inputs in the next level. For preventing oscillation, we do not optimize RP-GAN immediately once a single deconvolutional result is obtained. We set a buffer to store deconvolutional results of different levels produced by a frozen generator. Considering the results of the front levels are more influential than the latter ones, we make the results of the front levels occupy a larger proportion of the buffer. After getting a certain number of samples, we then renew and reinforce RP-GAN on the buffer by randomly feeding these samples. This strategy is somewhat like the widely-used experience replay method and fixed Q-targets in reinforcement learning. 

\begin{figure}[b]
\centering
  \includegraphics[scale = 0.30]{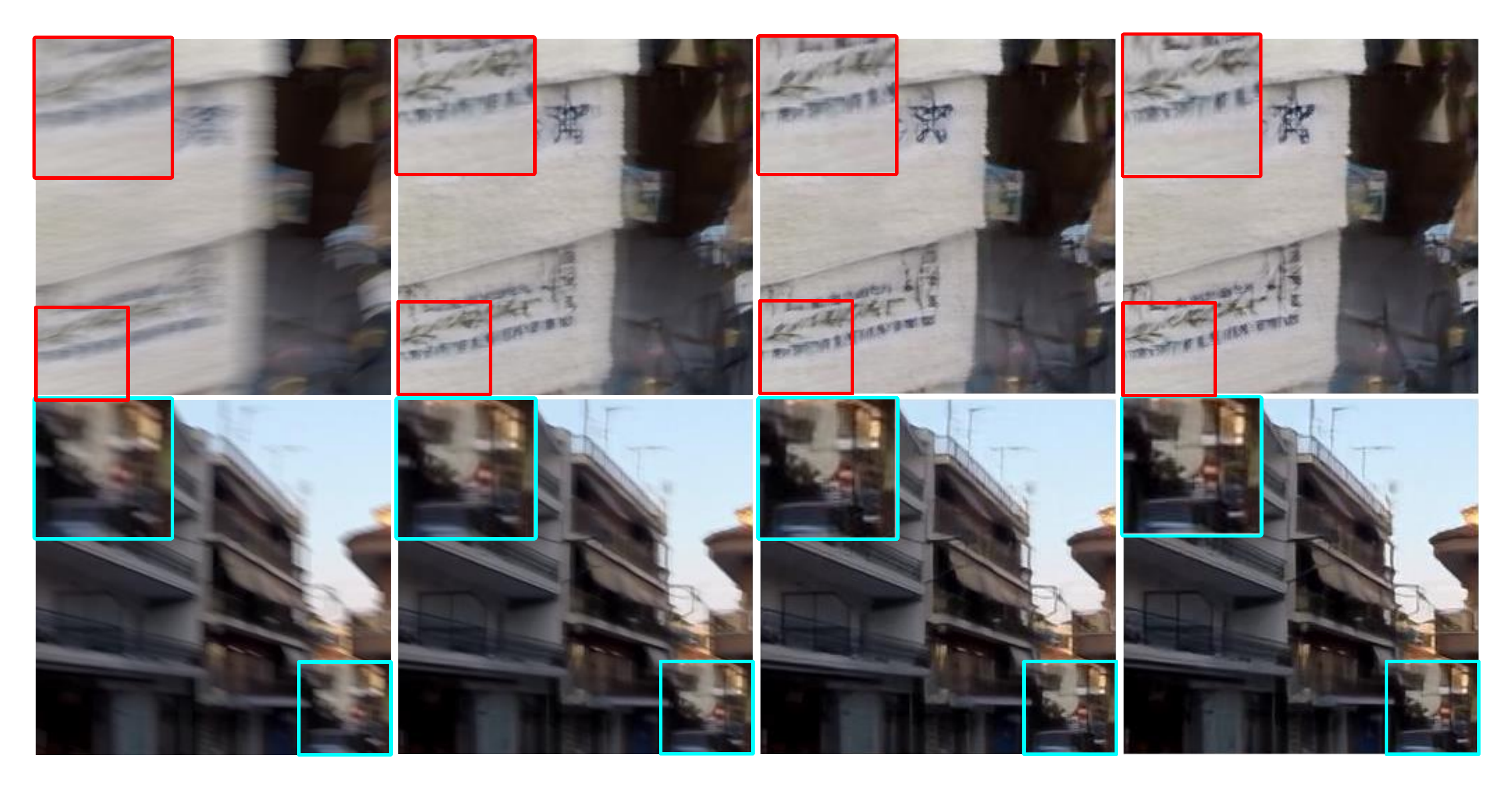}
  \caption{The images from left to right are blurry image, result of level 1, result of level 2, result of level 3 respectively. We see the RP-GAN outputs image from blurry to clear with recurrence going on.}
  \label{fig:inner}
\end{figure}

Another trouble occurred in our training is that, if we use the blur kernels from SEN, the results will show many artifacts even after sufficient training (we show these results in Appendix B). This is because the estimated blur kernels are not completely correct. Even tiny errors of the kernels will cause amplified artifacts. This problem could not be autonomously addressed through the network's supervised learning because once network outputs the images with remained artifacts, the artifacts will then escalate in the next deconvolution process and  lead the inputs of the next level become even worse. The artifacts' corruption will then be gradually aggravated in the process of iteration. The way we solve that is forcing the intermediate results to be optimized from blurry to clear but not from artifacts-corrupted to clear. Specifically, we add an extra artifacts regularization term with the form of discriminator in RP-GAN, called artifacts-penalized discriminator. The discriminator regards blurred image and artifacts-corrupted image as two extremes. It imposes extra punishment to the generator when the generated image is closer to the artifacts end while imposes no punishment when it is closer to the blur end. The blurry content is generated by feeding the original blurred image to the generator for compensating the artifacts. Since the deconvolution model will trust the blurry intermediate result more in the next iteration (as parameter $\beta$ has increased), it will produce fewer artifacts than the previous input, so that the image will become clearer and clearer after iterations. We show the variety of intermediate results in Figure \ref{fig:inner}.  Next we will look at the generator and the discriminators in detail.
\subsubsection{Generator}
%Several works build neural network to train an implicit prior for non-blind restoration \cite{Schuler2013A,Zhang2017Learning}. However, getting accurate blur kernel is challenging. Tiny blur-kernel-estimation mistake causes amplified artifacts. Thus, it is important to ensure that the network is robust to the mistakenly estimated blur kernel in blind deblurring. 
The inputs of the generator is the deconvolutional image: \(I^{*}\) and the concatenated observed blurred image: \(O\), which is used to compensate the heavily corrupted information in \(I^{*}\). We also skip connect the concatenated inputs with the last layer of the network to make the network to be more dependent on the offered information, thus to be sensitive to different inputs in different iterations.
\subsubsection{Discriminator}
Two discriminators are built to ensure RP-GAN to recover images from blurry to clear. cGAN discriminator plays a minimax game with \(G\), to penalize the difference between generated images and sharp images. Artifacts-penalized discriminator gives extra penalty to the artifacts. Two discriminators help \(G\) restore the clear content with the fewest artifacts.\\
\textbf{cGAN discriminator}
Since we have used the blurred image as  additional information, we build our generative and adversarial strategy following conditional generative and adversarial net \cite{cgan}. The objective of \(G\) and \(D\) in it can be expressed as:
\begin{equation}
\begin{split}
\arg \min\limits_{G}\max\limits_{D_{cGAN}} \quad  &\mathbb{E}_{I_{s},O}[log(D_{cGAN}\,(I_{s}\space | \space O))] + \\ &\mathbb{E}_{I^{*},O} [log(1-D_{cGAN}\,(G(I^{*}\space | \space O)\space | \space O)] 
\end{split}
\end{equation}
where \(I_{s}\) denotes the latent sharp image. \(O\) denotes the observed blurry image. \(I^{*}\) denotes the corrupted deconvolutional image. Generator \(G\) learns to minimize the objective against discriminator \({D_{cGAN}}\), which tries to maximize it. The generator is expected to generate images closer to natural ones through playing the minimax game.
The generated image will be sent to the discriminator with concatenated conditional variable. By comparing the blurred image with the generated image,  cGAN discriminator can better distinguish the generated one from the sharp one.\\
\noindent\textbf{Artifacts-penalized discriminator}
% 1. ensure improvement.  2. not sensitive to the mistacken blur kernel 
For constraining artifacts, we build an extra penalty term. Artifacts-penalized discriminator penalizes \(G\) if generated image is closer to the corrupted input than the sharp image.  We adapt WGAN \cite{wgan} to our discriminator. WGAN discriminator is trained to approximate the Wasserstein distance, or called Earth-Mover distance, between the generated distribution and the real distribution. But here, we train the artifacts-penalized discriminator as a Wasserstein distance between the distribution of corrupted input and the distribution of observed blurred image. The objective of artifacts-penalized discriminator can be expressed as : 
\begin{equation}\label{equation:wgantrain}
W(P_{data},P_{o})\approx \max \limits_{||D_{p}||\leq 1} \space \mathbb{E}_{x\sim P_{data}}\space [D_{p}(x)] - \mathbb{E}_{y\sim P_{o}}\space [D_{p}(y)]
\end{equation}
where \(P_{data}\) is the distribution of the corrupted input data and \(P_{o}\) is the distribution of the observed blurred image. \(x\) and \(y\) are the samples from distribution \(P_{data}\) and distribution \(P_{o}\) respectively. \(W(P_{data},P_{o})\) denotes the Wasserstein between two distributions. \(D_{p}\) denotes the artifacts-penalized discriminator. \(||D_{p}||\leq 1\) denotes that \(D_{p}\) should be 1-Lipschitz function. The penalty to $G$ is expressed as:
\begin{equation}\label{wgan}
\mathcal{L}_{p} (G,D_{p})= (D_{p}(G(I^{*}|O))-D_{p}(I_{s}))_{+}
\end{equation}
\((  )_{+}\) denotes the positive part of the content. As shown in  Eqn. (\ref{equation:wgantrain}), artifacts-penalized discriminator is trained to output a larger value for artifacts-corrupted image and a smaller value for blurry image. Thus, as shown in Eqn. (\ref{wgan}), punishment is only imposed when \(D_{p}(G(I^{*}|O))\) is larger than \(D_{p}(I_{s})\), which denotes the generated image is closer to the corrupted input than the sharp image in learned measure.
\subsubsection{Loss function}
Total loss of RP-GAN is expressed as:
\begin{equation*}
\begin{split}
\mathcal{L}_{RP-GAN} =& \mathcal{L}_{c}\,(G) + \gamma \, \mathcal{L}_{cGAN}\space (G,D_{cGAN}\,) +\\ & \lambda \mathcal{L}_{P}\space\space(G,D_{p})
\end{split}
\end{equation*}
where \(\mathcal{L}_{c}\space(G)\) denotes the content loss. \( \mathcal{L}_{cGAN}\)  denotes the adversarial loss. \(\mathcal{L}_{P}\)   denotes extra penalty. \(\lambda\) and \(\gamma\) are weighting constants. The objective of RP-GAN is simultaneously minimizing the three items until convergence.\\
\begin{figure}[H]
 \centering
  \includegraphics[width =0.5\textwidth]{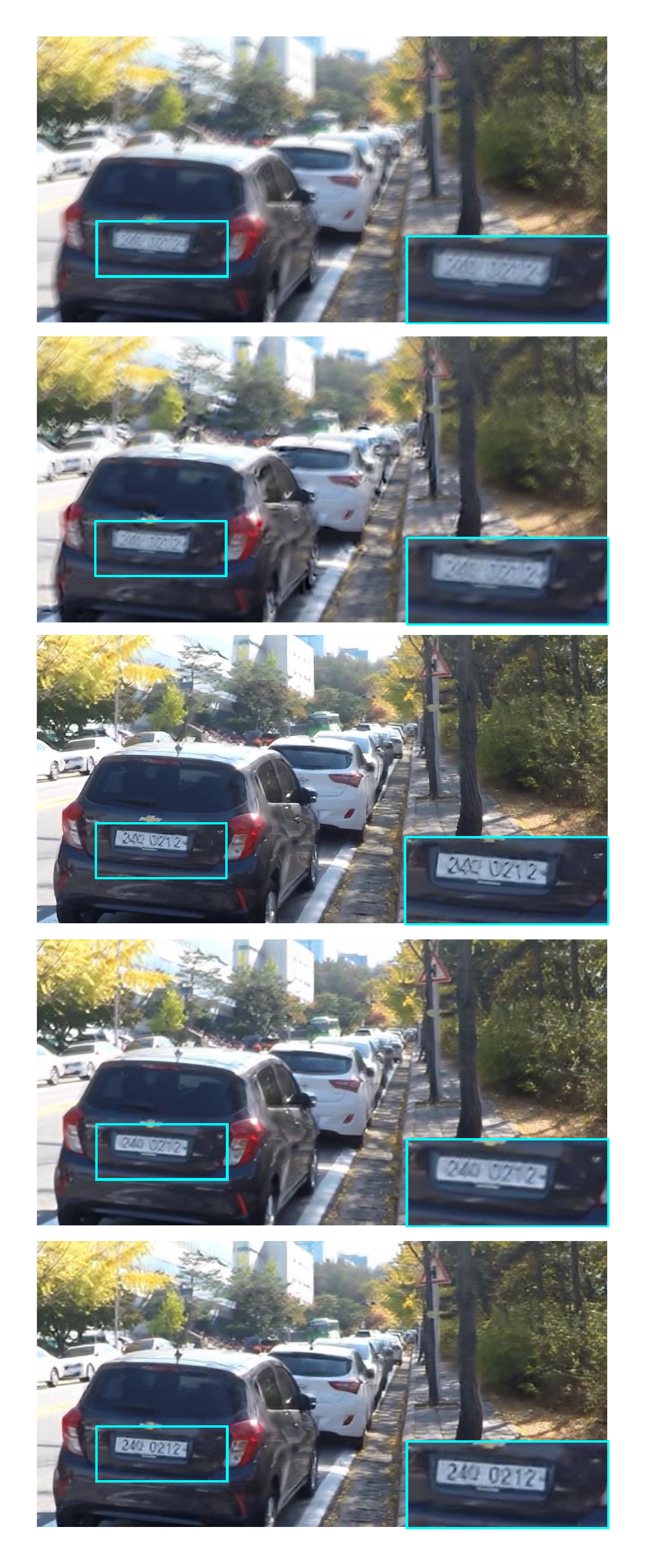}
  \caption{Test results on GOPRO dataset. Images from top to bottom are blurred image, results of Sun \textit{et al}. \cite{Sun2015Learning}, Nah \textit{et al}. \cite{Nah}, Tao \textit{et al}. \cite{Tao2018Scale} and ours respectively.}
   \medskip
  \label{fig:car}
\end{figure}
Specifically, adversarial loss can be expressed as:
\begin{equation*}
\mathcal{L}_{cGAN}\space (G,D_{cGAN}\,) = log(1-D_{cGAN}\;( G(I^{*}|O)|O))
\end{equation*}
Our content loss is perceptual loss \cite{Johnson2016Perceptual}, which can be
expressed as:
\begin{equation*}
\mathcal{L}_{c}\space \space(G) = \frac{1}{C*H*W}\sum^{C,H,W}_{c = 1,i = 1,j = 1}[\phi(G(I^{*},O))- \phi(I_{s})]
\end{equation*}
where \(\phi\) is a pre-trained network for abstracting the depth value of images. \(C,H,W\) are the channel, height and weight of a hidden layer of network \(\phi\). Perceptual loss compares the depth value of images instead of directly penalizing the difference of pixel-level information. In our experiment, it helps to prevent the network from over-fitting to the image contents and generate images with better perceptual quality. In practice, we compute the loss at layer relu2\_2 of the VGG16 network.
Extra penalty \(\mathcal{L}_{P}\)  is shown in Eqn. (\ref{wgan}).
\section{Global Iteration}
In experiment, we find sending the final result to the whole model again can further recover the residual blur. This effect is more apparent on more challenging datasets (GOPRO dataset \cite{Nah} and Kohler dataset \cite{K2012Recording}) than the simpler one (images generated by linear blur kernels). We speculate that this may be because after the blurry image first passes through our model, large and complex blur is segmented into small and simple residual blur which is easy to be accurately estimated and further handled. Although we can't fully explain it yet, we use it to improve our results. The quantitative comparison of the effect is shown in our experiment section.

%Note SEN could only produce nonuniform but linear motion flow maps. But in practice, the blur is more likely to be nonuniform and nonlinear. Since a nonliner blur kernel can be approximated by multiple linear blur kernels in different orientations, for processing the nonlinear blur images, we use the successive approximation strategy. Final results of RP-GAN will be sent to the system again for further processing the residue blur content. Through eliminating the approximated liner blur iteratively, the nonlinear blur can be gradually eliminated. As we estimate the blur kernels in pixel level, the global iterations have high efficiency. The effect of it is shown in Table \ref{table:gopro} and Table \ref{table:kohler}. The intact system for our blind deblurring is shown in Figure \ref{fig:system}.
%------------------------------------------------------------------------
\section{Experiments}
Our networks are implemented using PyTorch deep learning framework. All models are ran on NVIDIA 1080Ti GPU. 
%-------------------------------------------------------------------------
\subsection{Motion Blur Kernel Estimation}
\noindent
\textbf{Datasets}
Our data is generated following  \cite{Dong2017From}. For adapting to different scales of blur kernels, we generate two datasets. One is used to estimate small-scale blur kernels, the ceiling is set as \(v_{max} = u_{max} = 23\). The other is used to estimate large-scale ones, the ceiling is set as \(v_{max} = u_{max} = 46\).  Both datasets contain 12000 blur images generated from 1200 sharp images from Microsoft COCO \cite{coco} (Each sharp image generates 20 blur images). The 10000 images of the datasets are used for training. The other 2000 are used for testing. Two datasets are referred as MSCOCO-46 and MSCOCO-23 respectively.\\
\textbf{Training details}
SEN is trained by images cropped to 256*256 pixels. We set two stacks in practice. Both two stacks are trained by batch size 16. In the first stack, learning rate is set as \(1.25\,e^{-3}\)  in the first 50 epochs, and linearly decreases to \(1.25\,e^{-5}\) in the next 40 epochs. The second stack is trained 15 epochs by learning rate \(1.25\,e^{-5}\). Both of them use Adam solver with \(\beta_{1} = 0.9 \), \(\beta_{2} = 0.99\) and \(\epsilon = 10^{-8}\). \\
\textbf{Comparison}
We compare our method with  Sun \textit{et al}. \cite{Sun2015Learning} and Gong \textit{et al}. \cite{Dong2017From}. Sun \textit{et al}. estimated blur kernel in patch by CNN first, then further processed them with markov random field for smoothness. Gong \textit{et al}. proposed a FCN to estimate motion flow maps in pixel level and got previous state-of-the-art results. In our experiments, SEN shows higher qualitative results than previous networks by comparing mean square error (MSE). The comparison is shown in Table \ref{table:sen}.
\begin{table}  
\centering
\caption{MSE on motion flow estimation}
\scalebox{0.9}{
\begin{tabular}{|l|c|c|c|}  
\hline  
Dataset & Sun \textit{et al}.  & Gong \textit{et al}.  & Ours \\  
\hline  
 MSCOCO-23  & 24.34 & 5.54 & \textbf{2.71}\\  \hline  
MSCOCO-46  & 63.37 & 7.21 & \textbf{4.24}\\  
\hline  
\end{tabular}}
\label{table:sen}  
\end{table}  
%-------------------------------------------------------------------------
\subsection{Blind Deblurring}
For evaluating our blind deblurring method, we compare our method on mainstream benchmarks with previous state-of-the-art image deblurring approaches. PSNR and SSIM metrics are used to evaluate the restored image quality.\\
\textbf{RP-GAN training details}
RP-GAN is trained on two different datasets to ensure model's generalization. One contains 1000 synthetic blurred images generated following section 6.1. The other is  GOPRO dataset \cite{Nah}, which provides 2103 blur/clear pairs for training. All deconvolutional results are solved by the estimated blur kernels from the pre-trained SEN model. Batch size is set as 16. Learning rate is set as \(1e^{-4}\) for both generator and discriminator in the first 110 epochs, and linearly decreases to \(1e^{-6}\) in the next 80 epochs. We use the same solver as which used for SEN. Before the recurrence begin, we pre-train RP-GAN on the first level for 40 epochs. Then we set a buffer which stores 1000 images including 50\% first-level results, 30\% second-level results and 20\% third-level results. These images are randomly sent to RP-GAN for training 3 epochs before the buffer updated. The whole training process costs about 15 days.\\
\textbf{Comparisons on benchmarks}
We compare our algorithm with \cite{Sun2015Learning} \cite{Nah} and \cite{Tao2018Scale}. Sun \textit{et al}. \cite{Sun2015Learning}  learned blur kernel through CNN, then did non-blind deblurring by traditional deconvolution method. Nah \textit{et al}. \cite{Nah} and Tao \textit{et al}. \cite{Tao2018Scale} are two representative end-to-end learning methods for deblurring. Both of them used multi-scale strategy and generated perceptually convincing images. Tao\textit{ et al}. got previous state-of-the-art results on mainstream benchmarks. 

Methods are compared on two mainstream benchmarks, GOPRO dataset \cite{Nah} and Kohler dataset \cite{K2012Recording}. GOPRO dataset generated long-exposure blurry frames by averaging consecutive short-exposure frames from videos captured by high-speed cameras. The dataset provides 2103 clear/blur pairs for training and 1111 pairs for testing. The quantitative results on GOPRO testing dataset are listed in Table \ref{table:gopro}. Visual comparison is shown in Figure \ref{fig:car}. More comparisons are provided in Appendix B.
\begin{table}
\centering
\caption{Quantitative results on GOPRO testing dataset}
\scalebox{0.9}{
\begin{tabular}{|l|c|c|c|}
\hline
Method & Sun \textit{et al}.     & Nah \textit{et al}.    & Tao \textit{et al}.       \\ \hline
PSNR  & 24.64  & 29.08  & 30.10   \\ \hline
SSIM   & 0.8429 & 0.9135 & 0.9323  \\ \hline
Time   & 20min & 3.09s & \textbf{1.6s}  \\ \hline
Method & Ours.it1    &  Ours.it2    &  \textbf{Ours.it3}\\ \hline
PSNR  & 30.03  & 30.96  & \textbf{31.25}\\ \hline
SSIM & 0.9352 & 0.9487 & \textbf{0.9518}\\ \hline
Time &  17s & 32s & 47s\\ \hline
\end{tabular}}
\label{table:gopro}  
\end{table}
In Table \ref{table:gopro} and Table \ref{table:kohler}, it 1,2,3 denote the results of global iteration 1, 2 and 3 respectively. We see the results are improved with the iteration going on, and get the best result in the third iteration. Because containing the deconvolution process, our method will certainly cost more time than end-to-end learning methods. But using the deep neural network for optimization helps us take much less time than Sun \textit{et al}., who used a numerical method.
 \begin{figure*}[ht]
 \centering
  \includegraphics[scale = 0.34]{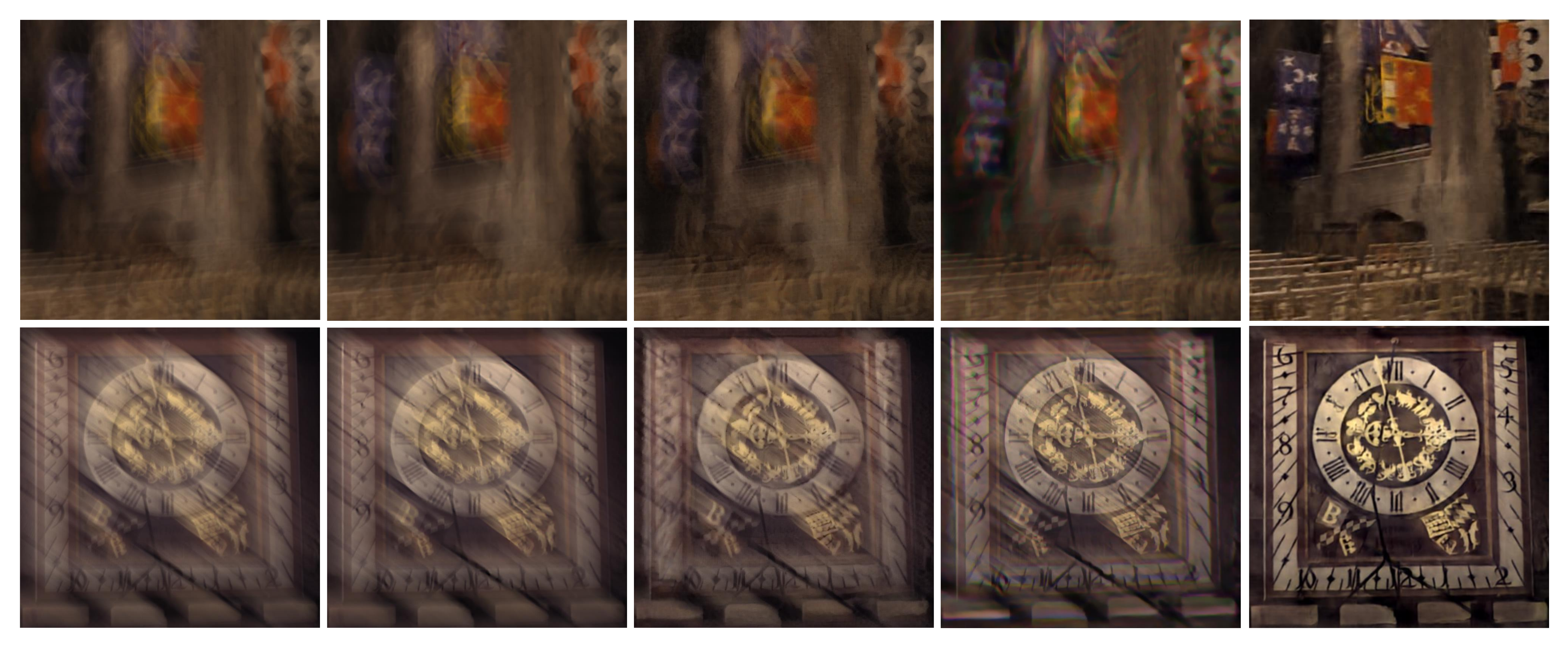}
  \caption{Test results on Kohler dataset. Images from left to right are blurred image, results of Sun \textit{et al}. \cite{Sun2015Learning}, Nah \textit{et al}. \cite{Nah}, Tao \textit{et al}. \cite{Tao2018Scale} and ours respectively.}
  \label{fig:kohler}
\end{figure*}
Kohler dataset \cite{K2012Recording} recorded and analyzed real camera motion, which is played back on a robot platform. The dataset consists of 4 images blurred with 12 different kernels for each of them. Comparisons on Kohler dataset are listed in Table \ref{table:kohler}. Visual comparison is shown in Figure \ref{fig:kohler}.
\begin{table}
\centering
\caption{Quantitative results on Kohler dataset}
\scalebox{0.9}{
\begin{tabular}{|l|c|c|c|}
\hline
Method & Sun \textit{et al}.    & Nah \textit{et al}.    & Tao \textit{et al}.   \\ \hline
PSNR    & 25.22  & 26.48  & 26.80   \\ \hline
SSIM   & 0.7735 & 0.8079 & 0.8375  \\ \hline
Method & Ours.it1 & Ours.it2 &\textbf{Ours.it3 }\\ \hline
PSNR & 27.25 & 28. 46 &  \textbf{29.19} \\ \hline
SSIM & 0.8642 & 0. 8933 & \textbf{0.9128} \\ \hline
\end{tabular}
 \label{table:kohler}}
\end{table}
\noindent
Comparing with end-to-end learning methods, our method restores clearer details on GOPRO test dataset. On Kohler dataset, our method outperforms others by a large margin. That is because the restoration ability of end-to-end learning based methods is only dependent on training dataset. It results in their failure of restoring extreme motions. This fault seems not so apparent on GOPRO testing dataset. But sometimes it restores unreasonable objects when the blur content is heavy. Such as Nah \textit{et al}. restore the slant license plate number in Figure \ref{fig:car}. Unlike the end-to-end learning methods, we build two networks independently estimate the motion flow and restore the deconvolutional results  in the optimization framework. Because of  learning a easier subtask, SEN can have a better awareness of motion than the end-to-end deblurring networks. Then RP-GAN can utilize the known motions to restore more reasonable objects. A typical example is shown in Figure \ref{fig:motion}.
On the other hand, end-to-end learning based methods perform worse on challenging blur images in Kohler dataset, which the motion is larger or more complex than which in GOPRO dataset. That's because their networks haven't 'seen' this pattern of motion. But since the deblurring framework we use is universally applicable, our method shows stronger generalization capability. 
\\
\textbf{ Recurrent strategy}
The number of the recurrent levels may also be a factor influencing the result. We compared the effect of setting different levels. The quantitative results of each setting are listed in Table \ref{table:recurrent}. We see the results improve a lot in the first three levels, and converges later. For the best efficiency, we take three levels in the paper.
\begin{table}
\centering
\caption{Comparison of different recurrent levels}
\scalebox{0.9}{
\begin{tabular}{|l|c|c|c|c|c|c|}
\hline
\multicolumn{2}{|l|}{Level} & 1    & 2    & \textbf{3 }   & 4    & 5    \\ \hline
\multicolumn{2}{|l|}{PSNR}  & 28.77  & 29.58  & \textbf{29.86}  & 29.83  & 29.84  \\ \hline
\multicolumn{2}{|l|}{SSIM}  & 0.9273 & 0.9305 & \textbf{0.9314} & 0.9312 & 0.9312 \\ \hline
\end{tabular}}
\label{table:recurrent}
\end{table}
\\
\textbf{The effectiveness of artifacts-penalized discriminator}
RP-GAN was trained without artifacts-penalized discriminator at first. The inexact blur kernels cause the results show many artifacts.  Visual comparison is shown in Appendix B.\\
\textbf{Comparison on real blurred images}
The comparisons with other methods on real-captured images are provided in Appendix B.
 \begin{figure}[h]
\centering
  \includegraphics[scale = 0.18]{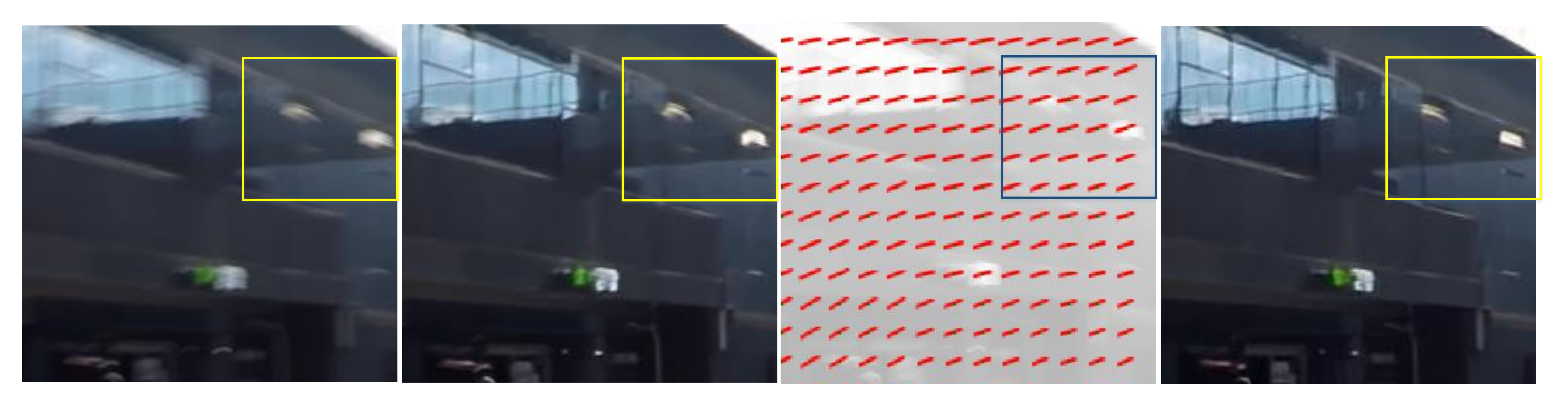}
 \caption{A typical example of the end-to-end learning methods' mistaken deblurring. Images from left to right is blurred image, result of Tao \textit{et al}. \cite{Tao2018Scale}, our estimated motion flow and our deblurred image respectively. We see Tao \textit{et al}. restore the vent to a weird shape. But since we have estimated the right motion in that location, we can restore it to a  reasonable shape.}
  \label{fig:motion}
\end{figure}
\section{Conclusion}
In this paper, we design the better neural networks to estimate the blur kernels and learn the implicit image prior, then further integrate the two parts to build an effective and convenient blind deblurring model. The experimental results show the effectiveness of the proposed method, which also proves that an optimization method based on theoretical guidance can overcome some shortcomings of  the end-to-end learning based methods and get better results. We believe there still have many possibilities worth exploring about this method. For example, estimating another form of the blur kernel, more experiments and analyses of global iteration, simplifying the deconvolution operation then jointly optimizing the estimation part and the optimization part, \textit{etc}. We are excited about these possibilities and would like to further investigate them in our future work.

%------------------------------------------------------------------
\newpage
{\small
\bibliographystyle{ieee}
\bibliography{egbib}

\begin{thebibliography}{10}\itemsep=-1pt

\bibitem{wgan}
M.~Arjovsky, S.~Chintala, and L.~Bottou.
\newblock Wasserstein gan.
\newblock 2017.

\bibitem{bigdeli2017image}
S.~A. Bigdeli and M.~Zwicker.
\newblock Image restoration using autoencoding priors.
\newblock {\em arXiv preprint arXiv:1703.09964}, 2017.

\bibitem{burger2012image}
H.~C. Burger, C.~J. Schuler, and S.~Harmeling.
\newblock Image denoising: Can plain neural networks compete with bm3d?
\newblock In {\em 2012 IEEE conference on computer vision and pattern
  recognition}, pages 2392--2399. IEEE, 2012.

\bibitem{dong2018denoising}
W.~Dong, P.~Wang, W.~Yin, and G.~Shi.
\newblock Denoising prior driven deep neural network for image restoration.
\newblock {\em IEEE transactions on pattern analysis and machine intelligence},
  2018.

\bibitem{Fergus2006Removing}
R.~Fergus, B.~Singh, A.~Hertzmann, S.~T. Roweis, and W.~T. Freeman.
\newblock Removing camera shake from a single photograph.
\newblock {\em ACM Trans. Graph.}, 25(3):787--794, 2006.

\bibitem{Dong2017From}
D.~Gong, J.~Yang, L.~Liu, Y.~Zhang, I.~Reid, C.~Shen, A.~V.~D. Hengel, and
  Q.~Shi.
\newblock From motion blur to motion flow: A deep learning solution for
  removing heterogeneous motion blur.
\newblock 2017.

\bibitem{imagetoimage}
P.~Isola, J.~Y. Zhu, T.~Zhou, and A.~A. Efros.
\newblock Image-to-image translation with conditional adversarial networks.
\newblock pages 5967--5976, 2016.

\bibitem{Johnson2016Perceptual}
J.~Johnson, A.~Alahi, and L.~Fei-Fei.
\newblock Perceptual losses for real-time style transfer and super-resolution.
\newblock In {\em European Conference on Computer Vision}, 2016.

\bibitem{Krishnan2011Blind}
D.~Krishnan, T.~Tay, and R.~Fergus.
\newblock Blind deconvolution using a normalized sparsity measure.
\newblock In {\em Computer Vision and Pattern Recognition}, pages 233--240,
  2011.

\bibitem{DeblurGAN}
O.~Kupyn, V.~Budzan, M.~Mykhailych, D.~Mishkin, and J.~Matas.
\newblock Deblurgan: Blind motion deblurring using conditional adversarial
  networks.
\newblock 2017.

\bibitem{K2012Recording}
R.~Köhler, M.~Hirsch, B.~Mohler, B.~Schölkopf, and S.~Harmeling.
\newblock {\em Recording and Playback of Camera Shake: Benchmarking Blind
  Deconvolution with a Real-World Database}.
\newblock Springer Berlin Heidelberg, 2012.

\bibitem{er}
L.~Li, J.~Pan, W.~S. Lai, C.~Gao, N.~Sang, and M.~H. Yang.
\newblock Learning a discriminative prior for blind image deblurring.
\newblock 2018.

\bibitem{coco}
T.-Y. Lin, M.~Maire, S.~J. Belongie, L.~D. Bourdev, R.~B. Girshick, J.~Hays,
  P.~Perona, D.~Ramanan, P.~Dollár, and C.~L. Zitnick.
\newblock Microsoft coco: Common objects in context.
\newblock {\em CoRR}, abs/1405.0312, 2014.

\bibitem{huang23}
Z.~Liu, R.~A. Yeh, X.~Tang, Y.~Liu, and A.~Agarwala.
\newblock Video frame synthesis using deep voxel flow.
\newblock pages 4473--4481, 2017.

\bibitem{cgan}
M.~Mirza and S.~Osindero.
\newblock Conditional generative adversarial nets.
\newblock {\em Computer Science}, pages 2672--2680, 2014.

\bibitem{Nah}
S.~Nah, T.~H. Kim, and K.~M. Lee.
\newblock Deep multi-scale convolutional neural network for dynamic scene
  deblurring.
\newblock pages 257--265, 2016.

\bibitem{hourglass}
A.~Newell, K.~Yang, and J.~Deng.
\newblock Stacked hourglass networks for human pose estimation.
\newblock pages 483--499, 2016.

\bibitem{Nimisha2017Blur}
T.~M. Nimisha, A.~K. Singh, and A.~N. Rajagopalan.
\newblock Blur-invariant deep learning for blind-deblurring.
\newblock In {\em IEEE International Conference on Computer Vision}, pages
  4762--4770, 2017.

\bibitem{darkchannel}
J.~Pan, D.~Sun, H.~Pfister, and M.~H. Yang.
\newblock Blind image deblurring using dark channel prior.
\newblock In {\em IEEE Conference on Computer Vision and Pattern Recognition},
  pages 1628--1636, 2016.

\bibitem{TV2}
D.~Perrone and P.~Favaro.
\newblock Total variation blind deconvolution: The devil is in the details.
\newblock In {\em IEEE Conference on Computer Vision and Pattern Recognition},
  pages 2909--2916, 2014.

\bibitem{Schuler2016Learning}
C.~Schuler, M.~Hirsch, S.~Harmeling, and B.~Scholkopf.
\newblock Learning to deblur.
\newblock {\em IEEE Transactions on Pattern Analysis \& Machine Intelligence},
  38(7):1439--1451, 2016.

\bibitem{Schuler2013A}
C.~J. Schuler, H.~C. Burger, S.~Harmeling, and B.~Scholkopf.
\newblock A machine learning approach for non-blind image deconvolution.
\newblock pages 1067--1074, 2013.

\bibitem{Sreehari2016Plug}
S.~Sreehari, S.~V. Venkatakrishnan, B.~Wohlberg, G.~T. Buzzard, L.~F. Drummy,
  J.~P. Simmons, and C.~A. Bouman.
\newblock Plug-and-play priors for bright field electron tomography and sparse
  interpolation.
\newblock {\em IEEE Transactions on Computational Imaging}, 2(4):408--423,
  2016.

\bibitem{huang31}
S.~Su, M.~Delbracio, J.~Wang, G.~Sapiro, W.~Heidrich, and O.~Wang.
\newblock Deep video deblurring.
\newblock 2016.

\bibitem{Sun2015Learning}
J.~Sun, W.~Cao, Z.~Xu, and J.~Ponce.
\newblock Learning a convolutional neural network for non-uniform motion blur
  removal.
\newblock (CVPR):769--777, 2015.

\bibitem{huang33}
X.~Tao, H.~Gao, R.~Liao, J.~Wang, and J.~Jia.
\newblock Detail-revealing deep video super-resolution.
\newblock pages 4482--4490, 2017.

\bibitem{tao2018srndeblur}
X.~Tao, H.~Gao, X.~Shen, J.~Wang, and J.~Jia.
\newblock Scale-recurrent network for deep image deblurring.
\newblock In {\em IEEE Conference on Computer Vision and Pattern Recognition
  (CVPR)}, 2018.

\bibitem{Tao2018Scale}
X.~Tao, H.~Gao, Y.~Wang, X.~Shen, J.~Wang, and J.~Jia.
\newblock Scale-recurrent network for deep image deblurring.
\newblock 2018.

\bibitem{TV1}
C.~TF and W.~CK.
\newblock Total variation blind deconvolution.
\newblock {\em IEEE Transactions on Image Processing}, 12(3):370--375, 1998.

\bibitem{Xu2013Unnatural}
L.~Xu, S.~Zheng, and J.~Jia.
\newblock Unnatural l0 sparse representation for natural image deblurring.
\newblock pages 1107--1114, 2013.

\bibitem{huang39}
N.~Xu, B.~Price, S.~Cohen, and T.~Huang.
\newblock Deep image matting.
\newblock 2017.

\bibitem{Yan2016Blind}
R.~Yan and L.~Shao.
\newblock Blind image blur estimation via deep learning.
\newblock {\em IEEE Transactions on Image Processing}, 25(4):1910--1921, 2016.

\bibitem{Zhang2016Learning}
J.~Zhang, J.~Pan, W.~S. Lai, R.~Lau, and M.~H. Yang.
\newblock Learning fully convolutional networks for iterative non-blind
  deconvolution.
\newblock pages 6969--6977, 2016.

\bibitem{Zhang2017Learning}
K.~Zhang, W.~Zuo, S.~Gu, and L.~Zhang.
\newblock Learning deep cnn denoiser prior for image restoration.
\newblock pages 2808--2817, 2017.

\end{thebibliography}
}

\clearpage
\begin{appendices}

\section{RP-GAN structure \& SEN structure}

We use \textbf{CBR} \(k*k*c:s\) denote the convolution layer with batch normalization and relu activation function, the convolution layer has kernel size \(k*k\) and \(c\) output channels applied with stride \(s\) , use\textbf{ Res1} \(c\) denote the first kind of  Residual Block which has \(c\) output channels, \textbf{Res2} \(c\) denote the second kind of  Residual Block which has \(c\) output channels, then has:
\\
\textbf{Res1}  \(c\) : \(input \to CBR \; 3*3*c:1 \to  CBR \;  3*3*c:1 + input \to output\) \\
\textbf{Res2}\; \(c\)  : \(input \to CBR \; 3*3*c/2:1\to CBR \;  3*3*c/2:1\to CBR \; 3*3*c:1+ input \; \divideontimes \; CBR \; 3*3*c:1 \to output\)\\
\textbf{RP-GAN}
\\
\textbf{Generator} : \(input \to CBR \; 7*7*64:1 \to  CBR \;  3*3*128:2 \to  CBR \;  3*3*256:2 \to Res1 \; 256 \to Res1 \; 256  \to Res1 \; 256 \to Res1 \; 256 \to Res1 \; 256 \to Res1 \; 256 \to Res1 \; 256 \to  CBR \;  3*3*256:2 \to CBR \;  3*3*128:2 \to CBR \; 7*7*64:1 + input \divideontimes Res2\; 3  \to output\)\\
\textbf{DcGAN} : \(input \to CBR \;  4*4*64:2 \to CBR \;  4*4*128:2\to CBR \;  4*4*256:2 \to CBR \;  4*4*512:2 \to CBR \;  4*4*512:1 \to CBR \;  4*4*1:1 \to sigmoid \to output\)\\
\textbf{Dp} : \(input \to CBR \;  4*4*64:2 \to CBR \;  4*4*128:2\to CBR \;  4*4*256:2\to CBR \;  4*4*512:2\to CBR \;  4*4*512:1\to CBR \;  4*4*1:1\to output\)\\
\begin{figure*}[hb]
\centering
  \includegraphics[width=\textwidth,height=\textheight,keepaspectratio]{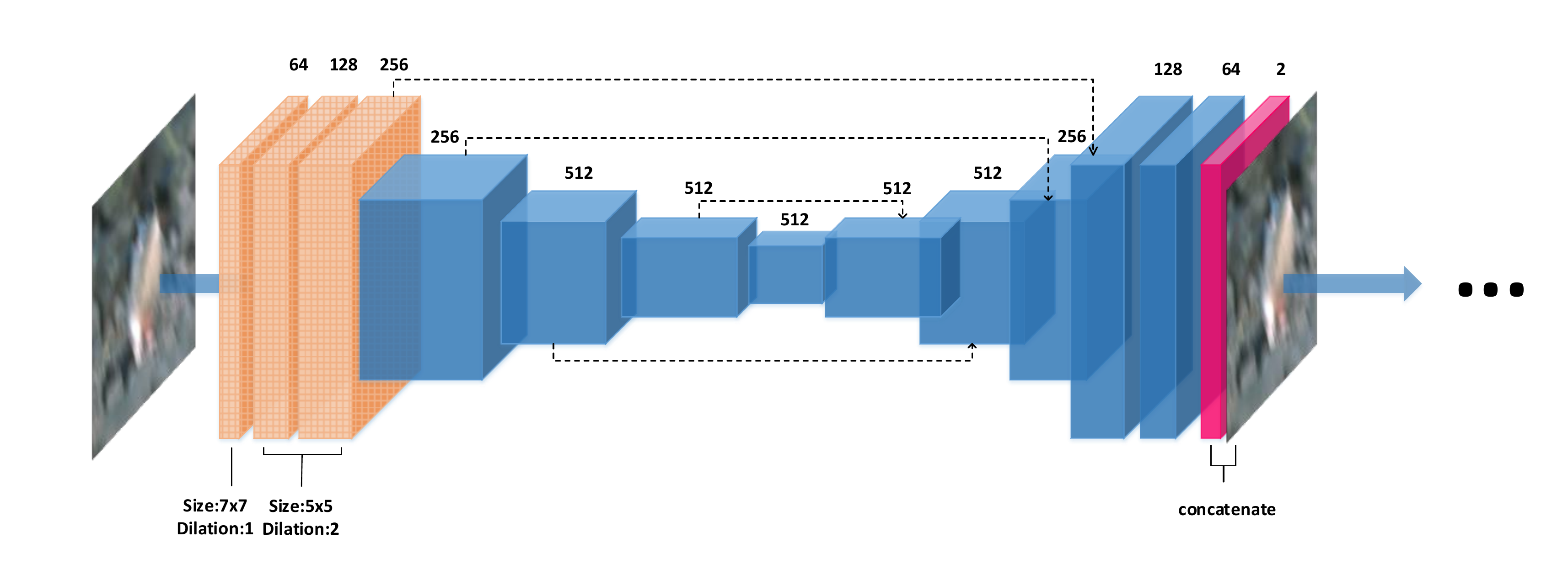}
   \caption{The first stack of SEN. Later stacks share the similar structure.}
   \medskip
\small
    \includegraphics[scale = 0.1]{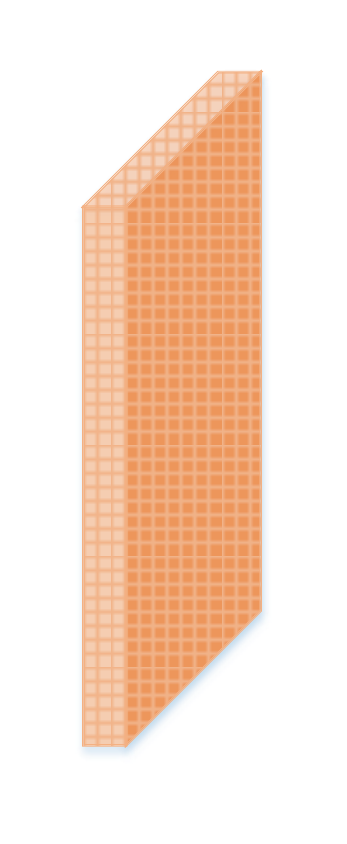} is dilated convolution. \includegraphics[scale = 0.1]{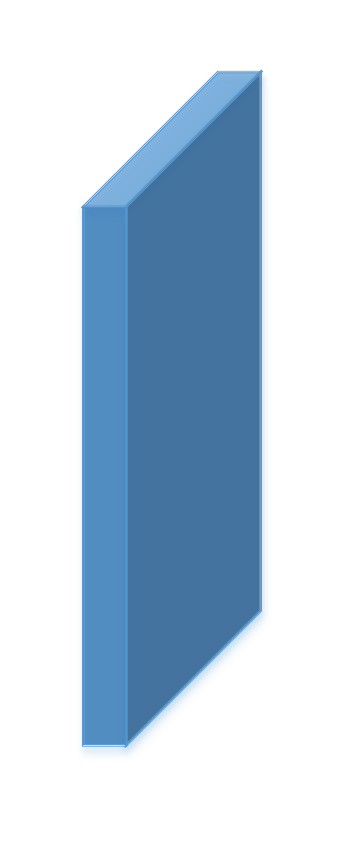} is residual learning block, \includegraphics[scale = 0.1]{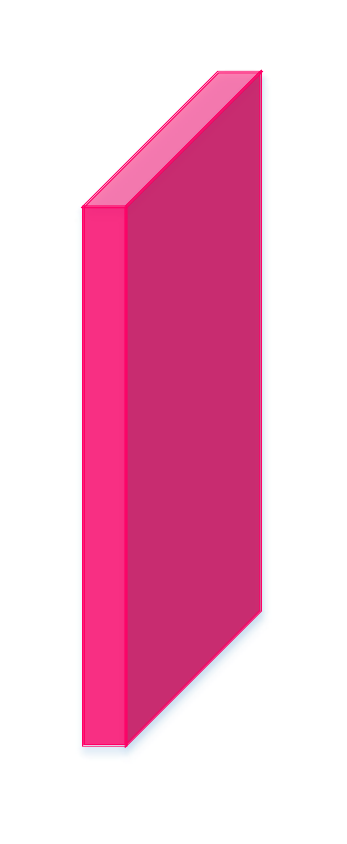} is objective motion flow map, \includegraphics[scale = 0.2]{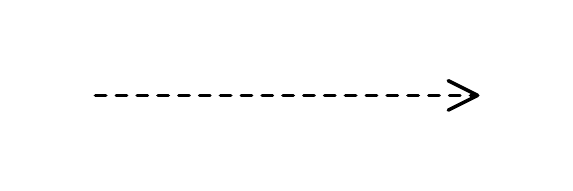} is skip connection.
  \label{fig:SEN structure}
\end{figure*}
  \textbf{SEN}\\
  \textbf{Stack 1} : \(input \to CBR \;  7*7*64:1 \to CBR \;  5*5*128:2\to CBR \;  5*5*256:2 \to maxpool \;  2*2 \to Res2 \;  256 \to maxpool \;  2*2 \to Res2 \; 512 \to maxpool \;  2*2 \to Res2 \;  512 \to maxpool \;  2*2 \to Res2 \;  512 \to upsample \;  2*2 \to Res2 \;  512 \to upsample \;  2*2 \to Res2 \;  512 \to upsample \;  2*2 \to Res2 \;  512  \to upsample \;  2*2 \to Res2 \;  256 \to upsample \;  2*2 \to Res2 \; 128 \to Res2 \; 64 \to Res2 \; 2 \to output\)\\
  The details of SEN is shown in Figure \ref{fig:SEN structure}.

 \clearpage
\section{Visual comparisons  }
\subsection{Comparison on GOPRO dataset}
\noindent
\begin{minipage}{1.0\textwidth}

  \centering
  \includegraphics[scale = 0.62]{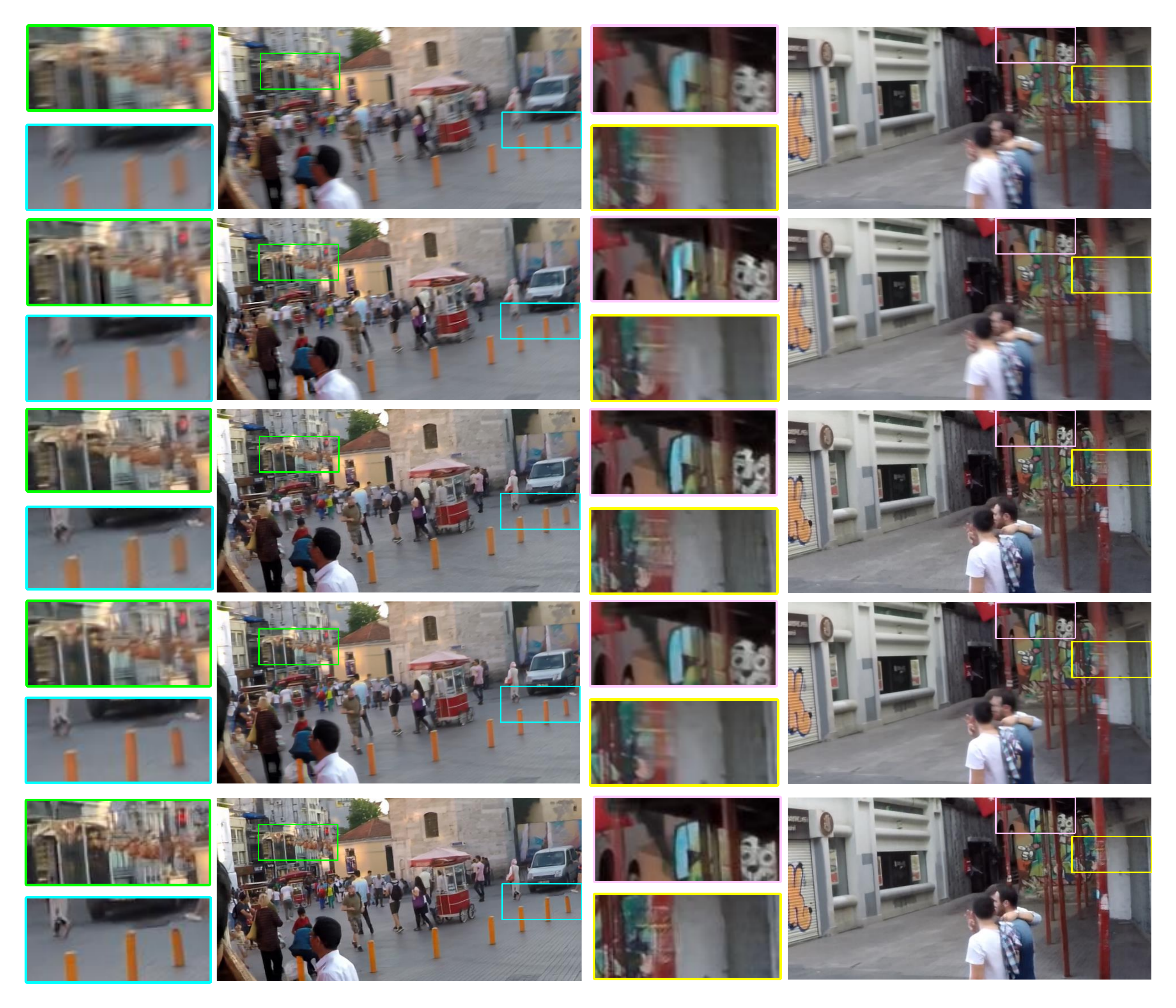}
  \captionof{figure}{More test results on GOPRO dataset. Images from top to bottom are blurred image, result of Sun \textit{et al}. \cite{Sun2015Learning},  Nah \textit{et al}. \cite{Nah}, Tao \textit{et al}. \cite{Tao2018Scale} and ours respectively.}\label{fig:figure1}
\end{minipage}

\clearpage
\subsection{The effectiveness of artifacts-penalize discriminator}
\noindent
\begin{minipage}{1.0\textwidth}

  \raggedleft
  \includegraphics[scale = 0.60]{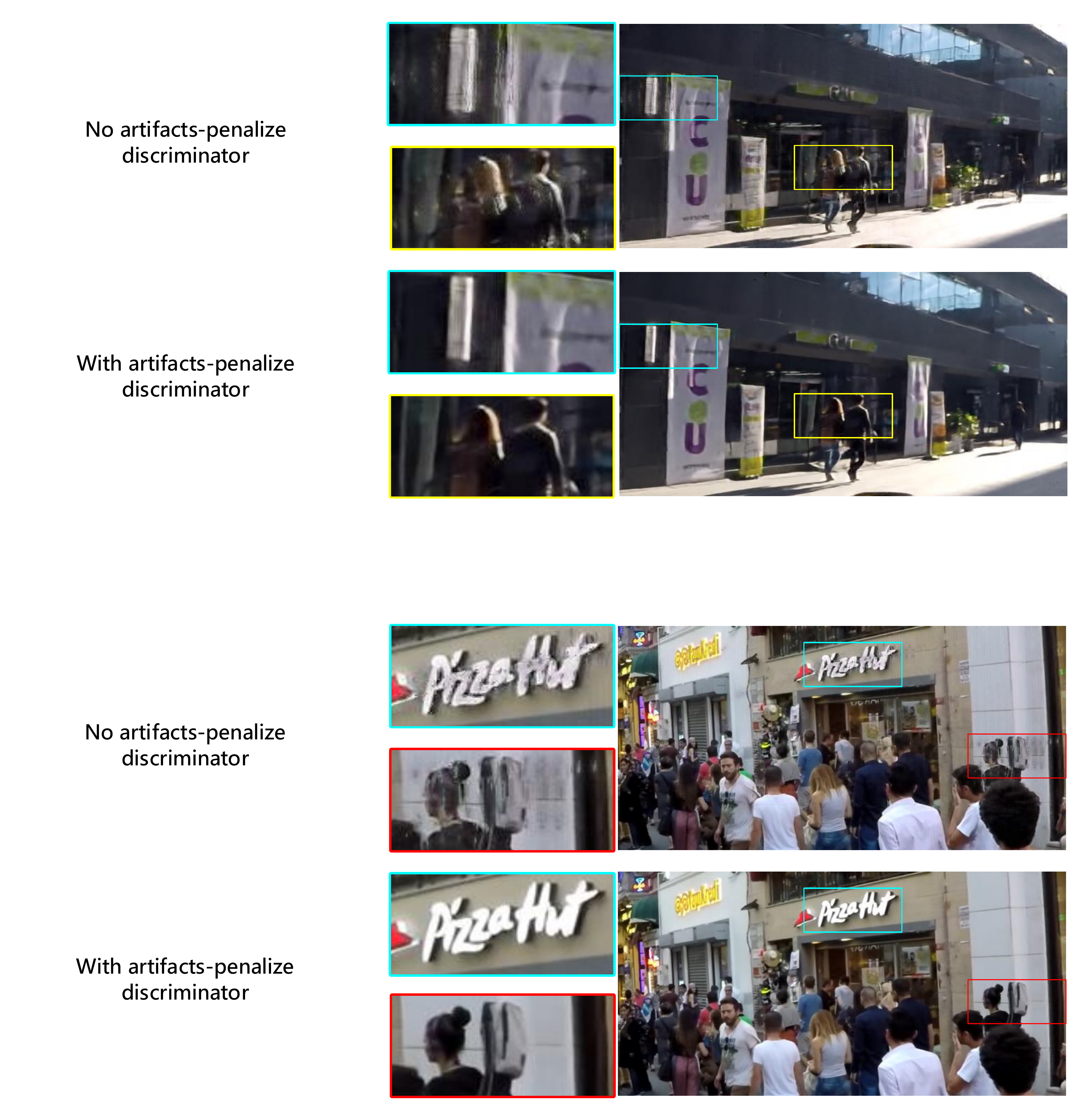}
  \captionof{figure}{Comparison of using artifacts-penalize discriminator before and after. Without artifacts-penalized discriminator, the quality of images with artifacts can't be improved through recurrence. Artifacts stay or even being worse in the end. }\label{fig:figure1}
\end{minipage}

\clearpage
\subsection{Comparison on real blurred images}
\noindent
\begin{minipage}{1.0\textwidth}

  \centering
  \includegraphics[scale = 0.60]{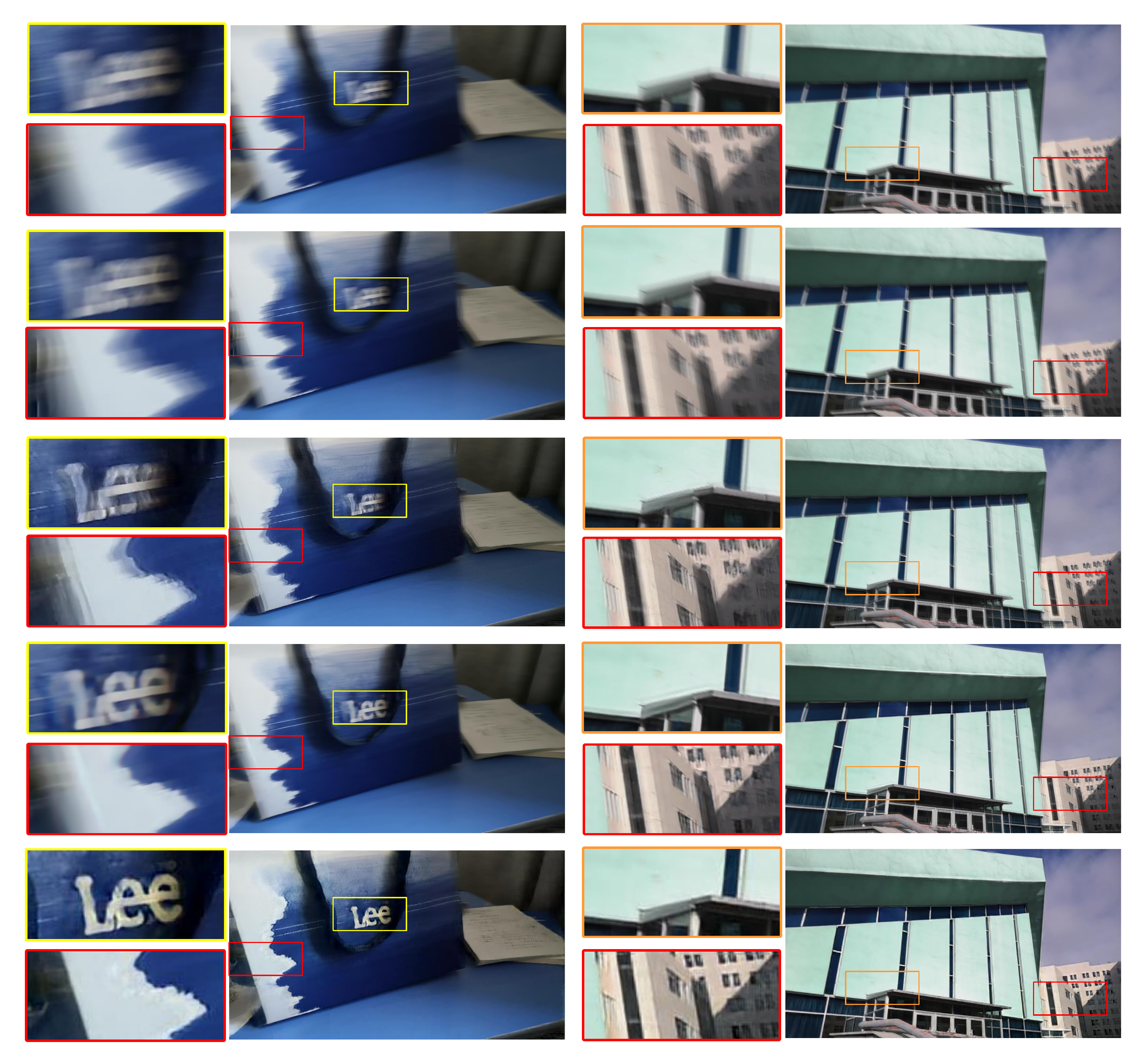}
  \captionof{figure}{Comparison on real blurred images. Images from top to bottom are blurred image, result of Sun \textit{et al}. \cite{Sun2015Learning},  Nah \textit{et al}. \cite{Nah}, Tao \textit{et al}. \cite{Tao2018Scale} and ours respectively. }\label{fig:figure1}
\end{minipage}

\clearpage

  \end{appendices}

\end{document}